\documentclass{statsoc}
\usepackage{graphicx, amsmath, amssymb, latexsym}
\usepackage{amsmath}
\usepackage{latexsym}
\usepackage{mathrsfs}
\usepackage{graphicx}
\usepackage{amsfonts}
\usepackage{amssymb}
\usepackage{natbib}
\usepackage{multirow}
\usepackage{ctable}
\usepackage{threeparttable}
\usepackage{rotating}
\usepackage{marvosym}
\usepackage{xcolor}
\usepackage[normalem]{ulem}

\usepackage{lineno}

\usepackage{soul} % For using \ul instead of \underline

\newcommand{\bel}{\begin{eqnarray}\label}
\newcommand{\eel}{\end{eqnarray}}
\newcommand{\bes}{\begin{eqnarray*}}
\newcommand{\ees}{\end{eqnarray*}}

\newcommand{\stkout}[1]{\ifmmode\text{\sout{\ensuremath{#1}}}\else\sout{#1}\fi}

\def\mathbold{\boldsymbol}

\def\bX{\mathbold{X}}
\def\bZ{\mathbold{Z}}
\def\bM{\mathbold{M}}
\def\bA{\mathbold{A}}
\def\bV{\mathbold{V}}
\def\bx{\mathbold{x}}

\def\btheta{\mathbold{\theta}}

\def\bz{\mathbold{z}}

\def\by{\mathbold{y}}

\def\bbeta{\mathbold{\beta}}

\def\ndef{\stackrel{\mathrm{d}}{=}}
\def\bb{\mathbold{b}}
\def\argmax{\mathrm{argmax}}
\def\argmin{\mathrm{argmin}}
\def\diag{\mathrm{diag}}
\def\model{\mathcal{A}}

\def\iden{\mathbold{\mathrm{1}}}

\def\bSigma{\mathbold{\Sigma}}

\def\bA{\mathbold{A}}
\def\bE{\mathbold{E}}

\def\ba{\mathbold{a}}
\definecolor{orange}{rgb}{1,0.5,0}

\title[Divide-and-Conquer]{DIVIDE-AND-CONQUER METHODS FOR BIG DATA ANALYSIS}
\author[X. Chen]{Xueying Chen}
\address{Novartis Pharmaceuticals Corp., 1 Health Plaza, East Hanover, New Jersey, USA}
\author[J. Cheng]{Jerry Q. Cheng}
\address{Department of Computer Science, New York Institute of Technology, New York, New York, USA}
\author[X. Chen, J.Cheng, and M. Xie]{Min-ge Xie}
\address{Department of Statistics, Rutgers University, Piscataway, New Jersey, USA}
\email{chenxuey@gmail.com}
\email{ jcheng18@nyit.edu}
\email{mxie@stat.rutgers.edu}

\begin{document}
\maketitle
\begin{abstract}
In the context of big data analysis, the divide-and-conquer methodology refers to a multiple-step process: first splitting a data set into several smaller ones; then analyzing each set separately; finally combining results from each analysis together. This approach is effective in handling large data sets that are unsuitable to be analyzed entirely by a single computer due to limits either from memory storage or computational time. The combined results will provide a statistical inference which is similar to the one from analyzing the entire data set.
This article reviews  some recently developments of  divide-and-conquer methods in a variety of settings, including combining based on parametric, semiparametric and nonparametric models,
online sequential updating methods, among others. Theoretical development on the efficiency of the divide-and-conquer methods is discussed. Examples of real-world data analyses are provided in various application areas.
\end{abstract}

%\keywords{
{{\bf Key Words} Big data, Combining Results from Independent Analyses, Distributed Computing, Divide-and-Conquer}

\section{Introduction}
With ever advancing computing and storage technologies, we frequently have access to large data sets gathered from a variety of sources with information sensing or collecting capabilities, such as the Internet of Things devices (e.g., mobile devices), social media, consumer activities, etc. As a result, we are facing information explosions in the era of big data, which has imposed both opportunities and challenges. Extremely large data sets usually not only have high-dimensional exploratory variables, but also large sample sizes. Often it is impossible to analyze entirely or even store such data in a single computer.
To solve this problem, tremendous efforts have been invested to reduce computational difficulties with various models.
Among them, the divide-and-conquer approach was proposed  and has been widely adopted as a general framework to handle and analyze extremely large data sets. Under this framework, we divide an original problem into several sub-problems, analyze them separately, and then combine the results to provide an inference.

The divide-and-conquer approach is easy to implement. In the \emph{divide} step, one could randomly split an entire data set into several subsets. Sometimes big data sets are available in separate chunks by nature. For example, they might be stored in multiple machines due to the storage limit of one machine, or generated continuously as streaming data.
Next  each subset is analyzed separately to provide statistical results  with no extra or less model fitting or programming efforts. In the \emph{combine} step, a simple and straightforward method is to average (with or without weights) the results from the subset analyses to obtain estimators. This combining strategy has been applied in various models such as  regularized estimators in generalized linear models, M-estimators as well as functional estimators in kernel regression models and non-parametric additive models.

The divide-and-conquer methodology can be described in a general term as follows (cf., e.g., \citet{chen2018quantile}). When an entire data set is split into $K$ subsets, $\{\mathcal{D}_1, \dots,\mathcal{D}_K\}$, a low dimension statistic $T_k=g_k(\mathcal{D}_k)$ is constructed for each subset with some function $g_k(\cdot)$. Then the final result is obtained by aggregating $\{T_k, k=1,\dots, K\}$ through an aggregation function $G(\cdot)$. In many studies, $g_k(\cdot)$ is selected as the same estimation function or test statistics for the entire data which saves efforts of additional model fitting.
It may still be computationally intensive to analyze smaller data sets $K$ times with some complex models.

Note that, if the simple average is used as the aggregation function, biases may not be reduced  because each subset now has a much smaller sample size. Some recent works, e.g.,  \citet{chen2018quantile, wang2018fast, jordan2019communication, wang2017efficient, fan2019communication}, have proposed to apply the divide-and-conquer algorithm iteratively with linearization of the original estimation function. It is worthwhile noting that only one subset would have $g_k(\cdot)$ the same as the estimation function of the entire data set in each iteration.
This multi-round divide-and-conquer algorithm could further reduce computational burden but require additional model fitting to obtain $g_k(\cdot)$ and $G(\cdot)$ functions.

% This article will review recent developments in divide-and-conquer algorithms.
The remaining article is organized as follows.
Section 2 reviews the
divide-and-conquer algorithms in
linear regression models, a simple case that can provide a key insight on the developments of the divide-and-conquer methodology. % where matrix factorization permits.
 Section 3 presents divide-and-conquer algorithms and their properties in parametric models, including sparse high dimensional models, M-estimators, Cox regression  and 	quantile regression models. Performance on
non-standard problems is also reviewed. Section 4 explained the algorithm in nonparametric and semiparametric models.
Section 5 describes divide-and-conquer applications in the setting of online sequential updating.
Section 6 discusses a method in small $n$ but large $p$ situation, where the division is vertically {\color{black}among} $p$ number of covariates.
Section 7 explains Bayesian divide-and-conquer and median-based combining methods. 
{\color{black}Section 8 provides some real data analyses using divide-and-conquer methods in various application areas.}
Section 9 concludes this paper with discussions.

% \section{Matrix factorization}
\section{Linear regression model}
We first consider a regular % Gaussian
linear regression model
$$y_i=\bx_i^T \bbeta + \epsilon_i, % \epsilon_i  \sim N(0, \sigma^2),
i=1,\ldots, n,$$
where $y_i$ is a response variable, $\bx_i$ is a vector of $p$ explanatory variables, $\bbeta$ is a vector of $p$ unknown parameters, and $\epsilon_i$ is an error noise of mean $0$.  %normally distributed with unknown standard error of $\sigma^2$.
The ordinary least square (OLS) estimator using the entire data is $\hat{\bbeta}=(\bX^T\bX)^{-1}\bX^T\by$, where $\by=(y_1,\dots,y_n)^T$ is the response vector, and $\bX=(\bx_1,\dots,\bx_n)^T$ is the design matrix with the assumption that $\bX^T\bX$ is invertible.

Suppose the entire data set is split into $K$ subsets, the $k^{th}$ subset has $n_k$ observations:$(\bX_k, \by_k)$, and the OLS estimator using the $k^{th}$ subset is $\hat{\bbeta}_k =(\bX_k^T\bX_k)^{-1}\bX_k^T\by_k$.   A combined estimator  \citep{lin2011aggregated,chen2014split} is proposed as
\bes
\hat{\bbeta}^{(c)}= (\sum_{k=1}^{K} \bX_k^T\bX_k)^{-1}\sum_{k=1}^K\bX_k^T\bX_k\hat{\bbeta}_k.
\ees

By simple algebra, we can show that the combined estimator is exactly identical to the whole-data solution:
\bel{ols}
\hat{\bbeta}^{(c)}=%\stkout{(\sum_{k=1}^{K} \bX_k^T\bX_k)^{-1}\sum_{k=1}^K\bX_k^T\bX_k\hat{\bbeta}_k=}
\Big(\sum_{k=1}^{K} \bX_k^T\bX_k \Big)^{-1}\sum_{k=1}^K\bX_k^T\by_k
=(\bX^T\bX)^{-1}\bX^T\by=\hat{\bbeta}
\eel

\citet{lin2011aggregated} extend the results of regular Gaussian linear regression models to the setting of general estimating equation (EE) estimation,
where the EE estimator $\hat{\bbeta}$ is the solution to the EE:  $\bM_n(\bbeta)=\sum_{i=1}^n \psi(\bx_i,y_i;\bbeta)=0$.
Since $\psi$ is often nonlinear, explicit analytical form of $\hat{\bbeta}$ is not available and thus combined estimator cannot be obtained straightforwardly. To solve this problem, \citet{lin2011aggregated}  approximate the nonlinear EE  by its first-order Taylor expansion at the EE estimator $\hat{\bbeta}_k$ based on the data of $k^{th}$ subset: ($\bx_{ki},y_{ki}: i=1,\ldots,n_i$):
\bel{ee}
\bM_{n_k}(\bbeta)=\{\bA_{k}(\hat{\bbeta}_k)\}(\bbeta-\hat{\bbeta}_k) + \mathbold{R}_k,
\eel
where $\bA_{k}(\bbeta)=-\sum_{i=1}^{n_k}\partial\psi(\bx_{ki},y_{ki};\bbeta)/\partial\bbeta$ and $\mathbold{R}_k$ is the remainder.
Now the combined EE estimator $\hat{\bbeta}_{k}$ is the solution to $\sum_{k=1}^K \{\bA_{k}(\hat{\bbeta}_k)\}(\bbeta-\hat{\bbeta}_k)=0$ when $\mathbold{R}_k$ is negligible.  \citet{lin2011aggregated} propose the combined estimator as:
\bel{combinedlin}
\hat{\bbeta}^{(c)}=\Big ( \sum_{k=1}^{K} \bA_k(\hat{\bbeta}_k) \Big )^{-1}\sum_{k=1}^K\{\bA_{k}(\hat{\bbeta}_k)\}\hat{\bbeta}_k.
\eel

\section{Parametric models}
\subsection{Sparse high dimensional models}
In supervised learning problems, both explanatory variables and response variables are observed. To handle extremely large scale data sets with enormous sample sizes and often possibly a huge number of the explanatory variables at the same time, regularized regression models have been studied in depth with various approaches proposed, such as LASSO estimator
\citep{tibshirani1996regression,chen2001atomic}, LARS algorithm \citep{efron2004least}, SCAD estimator \citep{fan2001variable} and
MCP estimators \citep{zhang2010nearly}. However, these approaches are computationally intensive and have cubic or higher algorithmic complexity in the sample size dimension \citep{chen2014split}. Divide-and-conquer strategy can substantially reduce computing time and computer memory requirement for such models.

Consider a generalized linear model.
Given explanatory variables $\bX=(\bx_1,\dots,\bx_n)^T$, the conditional distribution of response variable $\by=(y_1,\dots,y_n)^T$ is assumed to follow the canonical exponential distribution:
\bel{eq:likelihood}
 f(\by;\bX,\bbeta)= \prod_{i=1}^nf_0(y_i;\theta_i)=\prod_{i=1}^n\bigg\{c(y_i)\mathrm{exp}\bigg[\frac{y_i\theta_i-b(\theta_i)}{\phi}\bigg]\bigg\},
\eel
where $\theta_i=\bx_i^T\bbeta, i=1,\dots,n$ and $\phi$ is a nuisance dispersion parameter. The log-likelihood function $\log f(\by;\bX,\bbeta)$ is then given by
\bes
\ell(\bbeta;\by,\bX)= [\by^T\bX\bbeta-\iden^T\bb(\bX\bbeta)]/n,
\ees
where $\bb(\btheta)=(b(\theta_1),\dots,b(\theta_n))^T$ for $\btheta=(\theta_1,\dots,\theta_n)^T$ and the function $b(\cdot)$ is a smooth function with second
derivatives. When $p$ is large (or grows with $n$) and $\bbeta$ is sparse (i.e., many elements of $\bbeta$ are zero) with $s$ non-zero entries,
a regularized likelihood estimator is often used, which is defined as, in a general form,
\bel{penlike}
\hat{\bbeta}_{\lambda}=\argmax_{\bbeta}\left\{\ell(\bbeta;\by,\bX)/n - \rho(\bbeta;\lambda)\right\},
\eel
where $\rho$ is the penalty function with tuning parameter $\lambda$.

Under the divide-and-conquer framework, after the entire data set is split into $K$ subsets, the regularized estimator for the $k^{th}$ subset is
\bel{pen}
\hat{\bbeta}_k=\argmax_{\bbeta} \left\{\ell(\bbeta;\by_k,\bX_k)/n_k - \rho(\bbeta;\lambda_k)\right\},
\eel
where $\ell(\bbeta;\by_k,\bX_k)/n_k$ is the log-likelihood function for the $k^{th}$ subset with sample size $n_k$, and $\rho(\bbeta;\lambda_k)$ is the penalty function with tuning parameter $\lambda_k$. Note that, since each $\hat{\bbeta}_k$ is estimated from a different subset of data, $\{j: \hat{\beta}_{k,j}\not=0\}$ the set of selected variables (non-zero elements) of $\hat{\bbeta}_k$ can be different from one to another.

In order to obtain a combined estimator, \citet{chen2014split} used a majority voting method to obtain the set of selected variables of the combined estimator, i.e., $ \hat{\model}^{(c)}\ndef\bigg\{j: \sum_{k=1}^K\mathbold{\mathrm{I}}(\hat{\beta}_{k,j}\not=0)>w\bigg\}$, where $w\in[0, K)$ is a prespecified threshold and $\mathbold{\mathrm{I}}$ is the indicator function. Then the following weighted average of $\hat{\bbeta}_{k,\hat{\model}^{(c)}}$, $k=1,\dots,K$, is proposed as the combined estimator:
\bel{combinedchen}
\hat{\bbeta}^{(c)}\ndef \bA\,\left(\sum_{k=1}^K\bA^T \{ {\bX}_{k}^T\bSigma(\hat{\btheta}_k) {\bX}_{k} \} \bA\right)^{-1}\sum_{k=1}^K{\bA^T \{{\bX}_{k}^T\bSigma(\hat{\btheta}_k) {\bX}_{k} \} \bA}\hat{\bbeta}_{k,\hat{\model}^{(c)}},
\eel
where $\hat{\btheta}_k=\bX_k\hat{\bbeta}_k$, $\hat{\bbeta}_{k,\hat{\model}^{(c)}}$ is the sub-vector of $\hat{\bbeta}_{k}$ confined by the majority voting set $\hat{\model}^{(c)}$ and $\bSigma(\btheta)=\diag(\sigma(\theta_1),\dots,\sigma(\theta_n))$ with $\sigma(\theta)=\partial^2b(\theta)/\partial^2\theta$. Also, $\bE=\diag(v_1, \ldots, v_p)$ is the $p \times p$ voting matrix with $v_j= 1$ if $\sum_{k=1}^K \mathrm{I}(\hat{\beta}_{k,j}\not=0)>w$ and $0$ otherwise, and $\bA=\bE_{\hat{\model}^{(c)}}$ is the $p\times |\hat{\model}^{(c)}|$ selection matrix. Here, for any index subset $S$ of $\{1, \ldots, p\}$, $\bE_{S}$ stands for a $p\times |S|$ submatrix of $\bE$ formed by columns whose indices are in $S$.

\citet{chen2014split} show that the combined estimator $\hat{\bbeta}^{(c)}$ is sign-consistent under some regularity conditions and converges at the regular order of $O(\sqrt{s/n})$ under the $L_2$ norm. The combined estimator also obtains asymptotic normality with the same variance as the penalized estimator using the entire data.
For the selection of the number of splits $K$, \citet{chen2014split} find that a stronger constraint on the growth rate of $p$ would be imposed in order to detect the same signal strength as the corresponding complete data set analysis under the infinity norm.

Another strategy for combination is to debias or desparsify regularized estimators obtained from subsets, which has been adopted by \citet{lee2015communication}, \citet{battey2015distributed} and
\citet{tang2016method}.
Using LASSO estimators in linear regression for illustration, the debiased LASSO estimator by \citet{javanmard2014confidence} is
\bes
\hat{\bbeta}^{d} \ndef \hat{\bbeta}_{\lambda} + n^{-1}{(\bX^T\bX/n)}^{-}\bX^T(\by-\bX\hat{\bbeta}_{\lambda}),
\ees
where $\hat{\bbeta}_{\lambda}$ is regularized estimator defined in (\ref{penlike}) with $L_1$ norm penalty and ${(\bX^T\bX/n)}^{-}$ is an approximate inverse of $\bX^T\bX/n$.

Both \citet{lee2015communication} and \citet{battey2015distributed} propose the simple aggregated debiased LASSO estimator as the combined estimator:
 \bel{combinedlee}
\hat{\bbeta}^{(c)}\ndef \sum_{k=1}^K \hat{\bbeta}_k^{d}/K =K^{-1}\sum_{k=1}^K [\hat{\bbeta}_k +  \{(\bX_k^T\bX_k)/n_k\}^{-}\bX_k^T(\by_k-\bX_{k}\hat{\bbeta})],
\eel
where $\hat{\bbeta}_{k}^{d}$ is the debiased LASSO estimator for the $k^{th}$ subset with sample size $n_k$.
\citet{lee2015communication} show that with high probability when the rows of $\bX$ are independent subgaussian random vectors, the error of the aggregated debiased LASSO estimator in $L_{\infty}$ norm is $O(\sqrt{\log(p/n)})+ O(sK \log(p)/n)$. When $n$ is large enough, the latter term is negligible compared with the former term. Same results are obtained in \citet{battey2015distributed}. To further reduce the computing cost, \citet{lee2015communication}  use a single matrix $\hat{\mathbold{\Theta}}$ to replace all the terms $(\bX_k^T\bX_k/n_k)^{-}$, for $k=1,...,K$, which used to be solved for each subset $k$ and thus made it the most computational expensive step.  Following \citet{van2014asymptotically}, a common $\hat{\mathbold{\Theta}}$ is constructed by a node-wise regression on the explanatory variables.

\citet{battey2015distributed} also tackle hypothesis testing problems using divide-and-conquer in the framework of the Wald and Rao's score tests. Consider a test of size $\alpha$ of the null hypothesis for any coefficient, $H_0: \beta_j =\beta_j^{H}$ against the alternative, $H_1: \beta_j \not= \beta_j^{H}$, $j=1,..., p$. A divide-and-conquer Wald statistic is proposed:
\bel{test1}
\bar{S_n}=\sqrt{n}\sum_{k=1}^K(\hat{\bbeta}_{k,j}^d-\beta_j^{H})/(\bar{\sigma}\sqrt{\mathbold{b}_{k,j}^T\mathbold{b}_{k,j}}) ,
\eel
where $\bar{\sigma}$ is an estimator for the standard deviation of error based on $K$ subsets and $\mathbold{b}_{k,j}$ is the $j^{th}$ column of $(\bX_k^T\bX_k/n_k)^{-}\bX_k^T$ which can be obtained from the following optimization algorithm:
\bes
\mathbold{b}_{k,j}=\argmin_{\mathbold{b}}\mathbold{b}^T\mathbold{b}/n_k, s.t. \| \bX_k^T\mathbold{b}/n_k -\mathbold{e}_j\|_{\infty}\le \vartheta_1, \|\mathbold{b}\|_{\infty}\le \vartheta_2,
\ees
where $\mathbold{e}_j$ is a $p\times 1$ vector with the $j^{th}$ entry being 1 and the others being 0; $\vartheta_1$ and $\vartheta_2$ are tuning parameters. A simple proposal for $\bar{\sigma}$ is given by
\bes
\bar{\sigma}^2=K^{-1}\sum_{k=1}^Kn_k^{-1}\|\by_k-\bX_k^T\hat{\bbeta}^d_k\|_2^2.
\ees
Similarly, a simple average of the score estimators from $K$ subsets is proposed as the divide-and-conquer score statistic.

\citet{battey2015distributed} show that the limiting distribution of the divide-and-conquer estimator is asymptotically as efficient as the full sample estimator, i.e.
 $$\lim_{n\to\infty} \mathrm{Var}(\hat{\bbeta}^{(c)}_j)/\mathrm{Var}(\hat{\bbeta}^d_j) -1 =0,$$ $j=1,...,p$. Note that the hypothesis testing method is only developed for low dimensional parameters.

\citet{tang2016method} utilize confidence distributions \citep{xie2011confidence} to combine bias-corrected regularized estimators from subsets with the advantage that it provides a distribution estimator for various statistical inference, e.g. estimation or hypothesis testing, can be established straightforwardly. Particularly, in the setting of generalized linear model (\ref{eq:likelihood}) with LASSO penalty, asymptotic confidence density for each subset is constructed as:
\bel{condensity}
\hat{h}_{n_k}(\bbeta) \propto \mathrm{exp}[-(2\phi)^{-1}(\bbeta-\hat{\bbeta}^d_k)\{\bX_k^T\bSigma(\bX_k\hat{\bbeta}_k^d)\bX_k\}(\bbeta-\hat{\bbeta}_k^d))],
\eel
where $\bSigma(\bX_k\hat{\bbeta}_k)$ is the diagonal weight matrix based on the variance function of a generalized linear model as defined in (\ref{combinedchen}). Following \citet{liu2015multivariate}, $K$ confidence densities are combined to derive a combined estimator as the solution of $
\hat{\bbeta}^{(c)}$ as:
\bel{combinedtang}
\hat{\bbeta}^{(c)}
&\ndef& \argmax_{\bbeta} \mathrm{log}\Pi_{k=1}^K\hat{h}_{n_k}(\bbeta) \\
&=& \{\sum_{k=1}^K\bX_k^T\bSigma(\bX_k\hat{\bbeta}_k^d)\bX_k\}^{-1}\{\sum_{k=1}^K\bX_k^T\bSigma(\bX_k\hat{\bbeta}_k^d)\bX_k\hat{\bbeta}^d_k\}.
\eel

\citet{tang2016method} show that the combined estimator (\ref{combinedtang}) is asymptotically equally efficient as the estimator using the entire data. Note that both \citet{chen2014split} and \citet{tang2016method} have the combined estimator in the weighted average form while \citet{lee2015communication} and \citet{battey2015distributed} have simple average estimator as the combined estimator.

\subsection{Marginal proportional hazards model}
In the setting of multivariate survival analysis, \citet{Wang2019suv} apply a divide-and-combine approach  in the marginal proportional hazards model \citep{spiekerman1998marginal} and the shared frailty model \citep{gorfine2006prospective}. They use the similar combination estimator as Equation (\ref{combinedlin}) with three different weight structures for $\mathbf{A}_k$:  1) the minus second derivative of the log likelihood; 2) the inverse of variance-covariance matrix of the subset estimator; 3) the sample size. They prove that under mild regularity conditions, the divide-and-combine estimator is asymptotically equivalent to the full-data estimator.

\citet{Wang2019suv} also proposed a confidence distribution \citep{xie2013confidence} based regularization approach for the regularized estimator by minimizing the following objective function:
\begin{equation}
Q(\boldsymbol{\beta}) = n (\boldsymbol{\beta}-\hat{\boldsymbol{\beta}}^{(c)})^T \hat{\boldsymbol{\Sigma}}_{c}^{-1} (\boldsymbol{\beta}-\hat{\boldsymbol{\beta}}^{(c)}) + n \sum_{j=1}^{d} \lambda_j |\beta_j|,
\end{equation}
where $\lambda_1, \lambda_2, \dots, \lambda_d$ denote the tuning parameters, and $|\cdot|$ is the absolute value of a scalar. %Since the penalty is only applied on the regression parameter $\boldsymbol{\beta} = (\beta_1, \beta_2, \dots, \beta_d)^T$.
With a proper choice of $\boldsymbol{\lambda} = (\lambda_1, \lambda_2, \dots, \lambda_d)^T$, the regularized estimator $\hat{\boldsymbol{\beta}}^{(c)}_\lambda$ has the selection consistency, estimation consistency, and an oracle property.

\subsection{One-step estimator and multi-round divide-and-conquer}
Consider the M-estimator for a parameter of interest $\theta$ obtained by maximizing empirical criterion function $m(x_i;\theta)$ of sample size $n$ and data $x_i, i=1,...,n$:
\bes
\hat{\theta}=\argmax_{\theta} \sum_{i=1}^n m(x_i;\theta).
\ees
When data is split into $K$ subsets, each subset is analyzed separately to provide an estimator $\hat{\theta}_k =\argmax_{\theta} M_k(\theta),$ where $M_k(\theta)=\sum_{i=1}^{n_k}m(x_{k,i};\theta)$ is the empirical criterion function of the $k^{th}$ subset with sample size $n_k$, for $k=1,...,K$.

\citet{shi2018massive} consider weighted average of estimators from subsets with weight depending on the subset sample size:
 \bes
\hat{\theta}^{(c)}=\sum_{k=1}^K\omega_k\hat{\theta}_k=\sum_{k=1}^K n_k^{2/3}\hat{\theta}_k/(\sum_{k=1}^K n_k^{2/3}).
\ees
They establish the asymptotic distribution of the combined estimator and show that the combined estimator converges at a faster rate and has asymptotic normal distribution if the number of subsets diverges at a proper rate as the sample size of each subset grows.

The aforementioned divide-and-conquer approaches all have the combined estimator in the form of either simple average or weighted average. To further enhance the performance of the combined estimator or reduce the computational burden of solving $K$ problems for complex models, one-step update approach has been developed. It basically utilizes Newton-Raphson update once or in iterations to obtain a final estimator.

For M-estimators, a simple average combined estimator is defined as:
\bes
\hat{\theta}^{(0)}=\sum_{k=1}^K\hat{\theta}_k/K.
\ees
On top of the simple average estimator for pooling, \citet{huang2015distributed} propose a one-step estimator $\hat{\theta}^{(1)}$ by performing a single Newton-Raphson update:
\bel{update}
\hat{\theta}^{(1)}=\hat{\theta}^{(0)}-[\ddot{M}(\hat{\theta}^{(0)})]^{-1}[\dot{M}(\hat{\theta}^{(0)}],
\eel
where $M(\theta)=\sum_{k=1}^KM_k(\theta)$, and $\dot{M}(\theta)$ and $\ddot{M}(\theta)$ are the gradient and Hessian of $M(\theta)$ respectively.
They show that the proposed one-step estimator has oracle asymptotic properties and has mean squared error of $O(n^{-1})$ under mild conditions. It is worth noting that the proposed method and results are only developed for low dimensional cases. Numerical examples show that the one-step estimator has better performance than simple average estimators in terms of mean square errors.

The strategy of one-step update is also used in sparse Cox regression models by \citet{wang2018fast} and quantile regression models by \citet{chen2018quantile} in addition to linearization of the original optimization problem.  Due to the complexity of these models, it takes a long time to solve the original problem for all subsets as well. Therefore, multi-round divide-and-conquer is proposed to further reduce computational burden. The idea is that the original problem is only solved once for {\it one} subset and its result is used to construct a statistic for every other subsets. Statistics from all subsets are aggregated.  This divide-and-conquer process is then repeated iteratively.

\citet{wang2018fast} propose to start with a standard estimator that maximizes the partial likelihood for Cox proportional hazards model of \textit{one} subset as an initial estimator. Then the initial estimator is updated iteratively using all subsets linearly, in the same form of (\ref{update}) with corresponding matrices, to approximate the maximum partial likelihood estimator without penalty. Lastly, the final penalized estimator is obtained by applying least square approximation to the partial likelihood function \citep{wang2007unified}, given the estimator obtained in the second step. Since the maximization of partial likelihood function is only solved once on a subset in the first step and the penalized estimator is based on linear approximation in the last step, computational time is reduced tremendously.

\citet{chen2018quantile} propose a divide-and-conquer LEQR (linear estimator for quantile regression) which has a similar scheme of \citet{wang2018fast}. Using the idea of smoothing, a linear estimator for quantile regression is developed given a consistent initial estimator. To apply the divide-and-conquer approach, an initial estimator is calculated based on \textit{one} subset using standard quantile regression method. Then corresponding weight matrices of all subsets are calculated and  aggregated to update the estimator by solving a linear system. The second step is then repeated iteratively to provide a final estimator. \citet{chen2018quantile} show that the divide-and-conquer LEQR achieves nearly the  optimal rate of the Bahadur remainder term and achieves the same asymptotic efficiency as the estimator obtained based on the entire data set.

\citet{jordan2019communication} develop a general framework called Communication-efficient Surrogate Likelihood (CSL) which starts with an initial value and gradients of the loss function are calculated for each subset at the initial value. Similarly, the loss function is simplified and linearized using Taylor expansion and gets updated from aggregated gradients from subsets. This process is repeated iteratively to provide a final result. \citet{jordan2019communication} illustrate this multi-round divide-and-conquer approach in regular parametric models, high dimensional penalized regression and Bayesian analysis. A similar approach for penalized regression models is developed by \citet{wang2017efficient} separately as well. For the multi-round divide-and-conquer approach, requirement for the number of splits or machines $K$ is much relaxed to $K \preceq \mathrm{poly}(n)$ in contrast of $K\ll n$ in one round divide-and-conquer approach.

Multi-round divide-and-conquer by \citet{wang2018fast} and \citet{chen2018quantile} rely heavily on good initiation that is already consistent due to the nature of Newton-type methods. The framework by \citet{jordan2019communication} and \citet{wang2017efficient} has no restriction on the initial value but still requires a moderate sample size of each subset.  \citet{fan2019communication} improve CSL by adding a strict convex quadratic regularization to the updating step and the regularization is adjusted according to the current solution during the iteration. This approach is called Communication-Efficient Accurate Statistical Estimators (CEASE) and can converge fast.

\subsection{Performance in non-standard problems}
In the setting of noisy matrix recovery, \citet{mackey2011divide} propose an algorithmatic divide-factor-combine framework for large scale matrix factorization. A matrix is partitioned into submatrices according to its rows or columns and each submatrix  can be factored using any standard factorization algorithm. Submatrix factorizatons are combined to obtain a final estimate by matrix projection or spectral reconstruction approximation. In the setting of noisy matrix factorization, consider matrix $\bf{M}=\bf{L_0}+ \bf{S_0}+\bf{Z_0} \in R^{m*n}$, where  a subset of $\bf{M}$ is available,  $\bf{L_0}$ has rank $r \ll m, n$, $\bf{S_0}$ represents a sparse matrix of outliers of arbitrary magnitude, and $\bf{Z_0}$ is a dense noise matrix. \citet{mackey2011divide} show that if $\bf{L_0}$'s singular vector is not too sparse or too correlated ($(\mu, r)-coherent$ condition) and entries of $\bf{M}$ are observed at locations sampled uniformly without replacement, divide-factor-combine algorithms can recover $\bf{L_0}$ with high probability.

\citet{banerjee2019divide} study the performance of the divide-and-conquer approach in non-standard problems where the rates of convergence are usually slower than $\sqrt{n}$ and the limit distribution is non-Gausian, specifically in the monotone regression setting. Consider $n$ i.i.d. observations $(y_i, x_i), i=1,\cdots, n$ from the model \bes
y_i=\mu(x_i)+\epsilon_i,
\ees
where $\mu$ is a continuous monotone (nonincreasing) function on $[0,1]$ that is continuously differentiable with $0<c<|\mu'(t)|<d<\infty$ for all $t\in[0,1]$; $x_i \sim \mathrm{uniform}(0,1)$ and independent of $\epsilon_i$ with mean $0$ and variance $v^2$. Let $\hat{\theta}$ denote the isotonic estimate of {\color{black}$\theta=\mu^{-1}(a)$} for any $a\in\mathbb{R}$. It is known that $n^{1/3}(\hat{\theta} - \theta)\to_d \tilde{\kappa}Z$, where  $Z$ is the Chernoff random variable and $\tilde{\kappa}>0$ is a constant.

If the entire dataset is split into $K$ subsets and each provides an estimator $\hat{\theta}_k, k=1,\cdots, K$. \citet{banerjee2019divide} shows that the simple average combined estimator $\hat{\theta}^{(c)}$ outperforms the isotonic regression estimator using the entire data when $K$ is a fixed integer:
\bes
\mathrm{E}[n^{2/3}(\hat{\theta}^{(c)}-\theta)^2] \to K^{-1/3}\mathrm{Var}(\tilde{\kappa}Z).
\ees
However, for a suitably chosen (large enough) class of models, i.e. a neighborhood of $\mu$, called $\mathcal{M}$, as the class of all continuous non-increasing functions that coincide with $\mu$ outside of $(x_0 -\varepsilon_0, x_0+\varepsilon_0)$ for some small $\varepsilon_0>0$, when $K\to\infty$,  \bes
\mathrm{lim}\mathrm{inf}_{n\to\infty} \mathrm{sup}_{\mathcal{M}}[n^{2/3}(\hat{\theta}^{(c)}-\theta)^2]=\infty,
\ees
whereas, for the estimator using the entire dataset,
\bes
\mathrm{lim}\mathrm{sup} _{n\to\infty}\mathrm{sup}_{\mathcal{M}}[n^{2/3}(\hat{\theta}^{(c)}-\theta)^2]<\infty.
\ees

%Consider $\btheta$ is a finite dimentional parameter of interest which can be estimated from i.i.d. data $\bx_i, i,\cdot, %n$ and $\bx_i$ has a common unknown distribution. Assume a natural estimator $\hat{\btheta}$ of $\btheta$ %converges to  a non-normal limit at a rate slower than $n^{1/2}$:
%\bes
%r_n(\hat{\btheta}-\btheta) \rightarrow_d G,
%\ees
%where $r_n=o(\sqrt(n))$ and $G$ is non-normal.

It indicates that the combined estimator, i.e. simple average of estimators obtained from subsets, outperforms the estimator using the entire data set in the sense of point-wise inference under any fixed model. The combined estimator converges faster than the estimator using the entire data set and is asymptotically normal. However, in over appropriately chosen classes of models, the performance of combined estimators worsens when the number of splits increases.

\section{Nonparametric and semi-parametric models}

Given a dataset $\{(x_i,y_i)\}_{i=1}^n$ consisting of $n$ i.i.d. samples drawn from an unknown distribution and the goal is to estimate the function that minimizes the mean-square error $E[(f(X)-Y)^2]$, where the expectation is taken jointly over $(X,Y)$ pairs {\color{black} and $X$ is a univariate random variable}.  Consider the kernel ridge regression estimator of the optimal function $f^*(x)\ndef E[Y | X=x]$:
\bel{krr}
\hat{f} \ndef \argmin_{f\in \mathcal{H}} \{n^{-1}\sum_{i=1}^n(f(x_i)-y_i)^2 +\lambda\|f\|_{\mathcal{H}}^2\},
\eel
where $\lambda$ is a tuning parameter and $\mathcal{H}$ is a reproducing kernel Hilbert space {\color{black} which is endowed with an inner product $<\cdot,\cdot>_{\mathcal{H}}$ and $\|f\|_{\mathcal{H}} = \sqrt{<f,f>_{\mathcal{H}}}$ is the norm in $\mathcal{H}$.}

\citet{zhang2015divide} propose  to split the entire dataset into $K$ subsets and for each subset calculate the local kernel ridge regression estimate $\hat{f}_k$, $k=1,...,K$ from (\ref{krr}) using only data from corresponding subsets.  The combined estimate is the average of local estimates:
\bel{combinedzhang}
\hat{f}^{(c)}=\sum_{k=1}^K\hat{f}_k/K.
\eel
\citet{zhang2015divide} establish the mean-squared error bounds for the combined estimate in the setting of $f^*\in \mathcal{H}$ as well as $f^*\notin \mathcal{H}$. They show that the combined estimate achieves the minimax rate of convergence over the underlying Hilbert space.

All approaches discussed so far in this article are developed in the context that homogenous data are observed, either stored in different machines or split into subsets. In the case that the entire data is already split into subsets and heterogeneity exits in different subsets, \citet{zhao2016partially} and \citet{wang2019additive} consider partially linear models.  Suppose we have data with $n$ observations $\{(y_i, \bx_i, \bz_i)\}_{i=1}^n$ and there are $K$ subpopulations and the $k^{th}$ subpopulation has $n_k$ observations: $(y_{k,i},\bx_{k,i}, \bz_{k,i})$, $i=1,\dots,n_k$.
\bel{aplm}
\by_{k}=\bX_k^T\bbeta_k +f(\bZ_k),
\eel
where $\by_k=(y_{k,1},\dots,y_{k,n_k})^T$, $\bX_k=(\bx_{k,1},\dots,\bx_{k,n_k})^T$ and $\bZ_k=(\bz_{k,1},\dots,\bz_{k,n_k})^T$. Here $f(\cdot)$ is common to all subpopulations. In this model, $\by_k$ depends on $\bX_k$ through a linear function that may vary across subsets and depends on $\bZ_k$ through a nonlinear function $f(\cdot)$ that is common to all subsets.

 \citet{wang2019additive} choose $f(\bZ_k)=\sum_{l=1}^L g_l(\bZ_k)$, $k=1,\ldots,K$, to be additive non-linear functions with $g_l(\cdot)$ as  unknown smooth functions estimated by the regression spline method. \citet{zhao2016partially} use the kernel ridge regression method to estimate function $f$. In both approaches, $\bbeta_k$ and $f$ are estimated based on each subset providing $\hat{\bbeta}_k$ and $\hat{f}_k$, $k=1,\ldots,K$. Since $\bbeta_k$ presents the heterogeneity among different subsets, no additional action is needed. Further combination is done for the commonality part by averaging to provide the final nonparametric estimate {\color{black} $\hat{f}=\sum_{k=1}^K\hat{f}_k/K$}. Both approaches can be applied to homogenous data as well, which can be handled with a divide-and-conquer approach.

\section{Online sequential updating}
For many divide-and-conquer approaches, it is assumed that all data are available at the same time although data may be stored in different machines or cannot be analyzed at once. However, in some applications, data may arrive in batches or in streams and exceed the capacity of a single machine for storage or analysis. The divide-and-conquer approach, generally referred as online sequential updating, can be extended to such cases.

In the case of ordinary least square estimator (\ref{ols}), suppose we have the weight matrix $\bV_{k-1}=\sum_{l=1}^{k-1}\bX_l^T\bX$ and the combined estimator $\hat{\bbeta}^{(c)}_{k-1}$ available using data from subsets $l=1,..., k-1$. Once data in the $k^{th}$ subset come in, the online estimator can be updated \st{to} \citep{schifano2016online} to:
\bel{combinedschifano}
\hat{\bbeta}_k^{(c)}=(\bX_k^T\bX_k +\bV_{k-1})^{-1}(\bX_k^T\bX_k\hat{\bbeta}_k + \bV_{k-1}\hat{\bbeta}_{k-1}^{(c)}),
\eel
where the initial values of $\hat{\bbeta}^{(c)}_0$ and $\bV_0$ are set to 0, and $\bV_k$ is updated to $\bV_k=\bV_{k-1}+\bX_k^T\bX_k$.

\citet{schifano2016online} also propose an online updating estimator for general EE estimators. Instead of performing Taylor expansion at the EE  estimator of the $k^{th}$ subset $\hat{\bbeta}_k$ (\citet{lin2011aggregated}), \citet{schifano2016online} consider an intermediary estimator:
\bes
\tilde{\bbeta}_k=\{\tilde{\bA}_{k-1}+\bA_k(\hat{\bbeta}_k)\}^{-1}[\sum_{l=1}^{k-1}\bA_k(\tilde{\bbeta}_l) \tilde{\bbeta}_k+ \{\bA_k(\hat{\bbeta}_k)\}\hat{\bbeta}_k],
\ees
where $\tilde{\bA}_{k-1}=\sum_{l=1}^{k-1}\bA_k(\tilde{\bbeta}_l)$ with $\bA_k(\bbeta)$ defined in (\ref{ee}) and the initial value of $\tilde{A}_0$, $\tilde{\bbeta}_0$ are set to 0.
Plug in $\tilde{\bbeta}_k$ to the first order Taylor expansion and by some algebra, one can obtain the online updating estimator as
\bes
\hat{\bbeta}^{(c)}_k=\{\tilde{\bA}_{k-1}+\bA(\tilde{\bbeta}_k)\}^{-1}\{\ba_{k-1}+\bA_{k}(\tilde{\bbeta}_k)\tilde{\bbeta}_k +\bb_{k-1} + \bM_{n_k}(\tilde{\bbeta}_k)\},
\ees
where $\ba_{k}=\sum_{l=1}^{k}\{\bA_k(\tilde{\bbeta}_k)\}\tilde{\bbeta}_k=\bA_{k}(\tilde{\bbeta}_k)\tilde{\bbeta}_k +\ba_{k-1}$ and $\bb_{k}=\sum_{l=1}^{k}M_{n_l}(\tilde{\bbeta}_l)=M_{n_k}(\tilde{\bbeta}_k)+\bb_{k-1}$ with initial values of $\ba_0=0$ and $\bb_0=0$.

\citet{wang2018online} address the online updating problem with emergence of new variables, i.e. new predictors become available midway through the data stream. Under the assumption that the true model contains these new variables, not only estimation of coefficients for newly available variables is needed, bias for previously existing variables should be corrected as well. The bias of existing variables for the online updating estimator $\hat{\bbeta}_{k-1}^{(c)}$ up to block $k-1$ can be corrected using data in block $k$ alone as the difference between OLS estimators with and without new variables. Then a weighted average similar to (\ref{combinedschifano})  is applied to update the cumulative estimator of existing variables, with extra care of the variance of a bias term. Estimate of new variables is based on data in block $k$ to start with. After that, updating for future blocks is a weighted average of full models.

\citet{kong2019efficiency} consider online updating for various kernel-based nonparametric estimators. They propose weighted sum updating:
\bes
\hat{f}_{k}(\bx) = (1-\alpha_k)\hat{f}_{k-1}(\bx) +\alpha_k K_{h_k}(\bx;\bX_k),
\ees
where $\alpha_k\in (0,1)$ is a pre-specified series of constants and $K_{h_k}$ is kernel function with bandwidth $h_k$. Note that the bandwidth $h_k$ is independent of previous observed data and only depends on new data $\bX_k$. They investigate the optimal choices of bandwidths and optimal choices of weights. Relative efficiencies of online estimation with regard to dimension $p$ are also examined.

\section{Splitting the number of covariates}

Under the sparse high dimensional setting in Section 3.1,  the divide-and-conquer approach would split a data set (of size $n$) into subsets of smaller sample size ($n_k$) where each data point has all the information available, i.e. response variable(s) and all explanatory variables. From a different perspective, \citet{song2015split} propose to split a high dimensional data set into several lower dimensional subsets, each of which has the same sample size as the entire data set but only a portion of explanatory variables. Furthermore, the  explanatory variables in subsets are mutually exclusive. Once data is split, Bayesian variable selection is performed for each subset based on marginal inclusion probability iteratively. Finally, variables selected from subsets are merged into a single set and another Bayesian variable selection is performed on the merged data set. This procedure is named as split-and-merge (SAM).

The proposed SAM method can reduce computational cost tremendously in ultrahigh dimensional settings where the number of explanatory variables is much larger than the sample size. This is because in the second step where the Bayesian variable selection is performed on the subsets, a great number of variables have been screened out. With extreme splitting where each subset only has one variable, SAM is similar to sure independence screening (SIS) \citep{fan2008sure}. However, unlike SIS which screens out uncorrelated explanatory variables individually, SAM utilizes joint information of all explanatory variables in a subset to filter explanatory variables which leads to more accurate selection. \citet{song2015split}  show that SAM can select true variables with non-zero coefficients correctly as the sample size becomes large.

\section{Bayesian divide-and-conquer and median-based combining}

\citet{minsker2014scalable, minsker2017robust} propose a robust posterior distribution in Bayesian analysis which also utilizes the divide-and-conquer scheme. Let $\pi$ be a prior distribution over the parameter space $\Theta$ and $\theta\in \Theta$. The entire sample is divided into $K$ disjoint subsets $\{\bX_k=(\bx_{k,1},\cdots,\bx_{k,n_k}), k=1,\cdots, K\}$.  Suppose $f_k(\theta |\bX_k, \pi)$ is the posterior distribution depending on subset $k$. \citet{minsker2017robust} defines the M-posterior as \bes
f^{(c)}(\theta | \bX_1,\cdots, \bX_K, \pi)= \mathrm{med} (f_1(\theta |\bX_1, \pi),\cdots, f_K(\theta |\bX_k, \pi)),
\ees
where the median is the {\it geometric median} defined for a probability measure $\mu$:
\bes
x_* = \mathrm{argmin}_{y\in \mathbb{Y}} \int_{\mathbb{Y}} (\|y-x\| - \|x\|)\mu(dx),
\ees
with $\mathbb{Y}$ be anormed space with norm $\|\cdot\|$ and $\mu$ be a probability measure on $(\mathbb{Y},\|\cdot\|)$ equipped with Borel $\sigma$-algebra.

Due to the property of a geometric median, there exists $\alpha_1\ge 0, \cdots,\alpha_K\ge 0$ and $\sum_{k=1}^K\alpha_k=1$ such that $f^{(c)}(\theta | \bX_1,\cdots, \bX_K, \pi) =\sum_{k=1}^K\alpha_k f_k(\theta |\bX_1, \pi)$ which leads to a weighted average of posterior distribution from subsets and the weights depend on the norm used on probability measure space. Note that it is possible to have $\alpha_k=1$ for one subset and the rest of weights are zero in which case the `median' is being selected as the combined posterior.

\citet{minsker2014scalable, minsker2017robust} further improve the robust posterior by replacing posterior distribution from subsets with {\it stochastic approximations}. The stochastic approximation can be obtained as a posterior distribution given each data point in a subset is observed $K$ times. \citet{minsker2017robust} show that the modified posterior yields credible sets with better coverage but $f^{(c)}(\theta | \bX_1,\cdots, \bX_K, \pi)$ often overestimates the uncertainty about $\theta$. Numerical algorithms to calculate the geometric mean of probability distributions are also provided.

%first order combining: median
The `median' based combing approach can be generalized to many other models, including non-Bayesian estimators.
\citet{minsker2019distributed} discuss that the averaging based combing approach attains the optimal converging rate if the bias of each subset estimator is small enough. However, if one or more subset estimators are deviating from norm, the combined estimator from averaging would be affected as well. Therefore, \citet{minsker2019distributed} propose to use a more robust combining approach such as median or a robust M-estimator and investigate the performance of median combined estimators. They demonstrate that the median combined estimator has a much slower converging rate if subset estimators remain the standard converging rate at regular conditions unless the number of subsets $K$ is limited and small. However, the converging rate can be improved with additional constraints when $K$ is as large as $O(\sqrt{n})$. Detailed investigations and discussions are illustrated for the median-of-mean estimators and maximum likelihood estimation.

Getting back to Bayesian divide-and-conquer, if the M-posterior by \citet{minsker2014scalable, minsker2017robust} combines posteriors from subsets through their median in the Wasserstein space of order one, \citet{srivastava2015wasp, srivastava2018scalable} combine the posteriors of subsets through the mean in the Wasserstein space of oder two which is called Wasserstein Posterior. They demonstrate that the proposed posterior converges in expectation and provide numerical algorithms for computation.

Bayesian divide-and-conquer approaches include the prior distribution in each subset's inference. In many approarches the prior is multiply-counted when the inference or posterior distribution is combined. But if the prior distribution is divided into pieces as well, e.g. fractional of prior ${\pi(\theta)}^{1/K}$ is used, it may be too weak to effectively regularize \citep{gelman2017expectation}. To solve this issue, \citet{gelman2017expectation} propose to use Expectation Propagation (EP) as a framework for Bayesian analysis in a distributed setting.  Expectation propogation (EP) is an iterative algorithm in which a target density $f(\theta)$ is approximated by a density $g(\theta)$ from some specified parametric family. The algorithm takes advantage of the natural factorization of likelihood function and the fact that the posterior distribution is proportional to the product of prior distribution and likelihood function: \bes
f(\theta)\propto \prod_{k=0}^K f_k(\theta),
\ees
where $f_k(\theta)$ is the likelihood function for subset $k, k=1,\cdots,K$ and $f_0(\theta)$ is the prior distribution.  Then the iterative algorithm is applied treating the prior distribution and likelihood functions equally. \citet{gelman2017expectation} review the general EP algorithm and provide its implementation for various Bayesian models as well.

\section{Real-world applications}
With the emerging of big data in different fields, the divide-and-conquer approach has a wide range of % real world 
applications as demonstrated in many articles.

Advances in genetics and molecular biology has dramatically increased our ability to collect massive data such as gene expressions or structures of chemical compounds. Questions such as relationships between phenotypes and candidate genes or screening of chemical compounds often arise. \citet{milanzi2014permutational} quantified expert opinions to assess 22,015 clusters of chemical compounds to identify those for further screening and development. \citet{meng2017effective} analyzed an Illumina HiSeq data set downloaded from the Cancer Genome Atlas (TCGA) Program (http://cancergenome.nih.gov) for 59 cancer patients with 20,529 genes using linear regression models. \citet{song2015split} illustrated the Bayesian split-and-merge methods in a metabolic quantitative trait loci experiment, which links SNPs data to metabolomics data as well as a polymerase chain reaction data set which contains 60 mouse samples of 22,575 genes' expression levels.

Divide-and-conquer approach has also been applied in social sciences and civil applications such as  the General Society Survey (GSS)  (http//gss.norc.org) which has collected responses about evolution and the growing complexity of American society since 1972 with approximately 28,000 respondents \citep{minsker2017robust}, the airline on-time performance data from the 2009 ASA Data Expo that includes flight arrival and departure details for all commercial flights within the U.S. from October 1987 to April 2008 \citep{schifano2016online, wang2016statistical}, manifest data which is compiled from custom forms submitted by merchants or shipping companies from the US custom offices and the Department of Homeland Security (DHS) \citep{chen2014split}, etc.

Online recommendation services of advertisements or news articles have received extensive attentions and massive data can be easily collected via internet. Different large scale advertisement datasets have been studied using the divided-and-conquer approach, e.g. a public advertisement dataset released by Criteo, which has 15 million instances with a binary outcome \citep{tang2018learning} and a Yahoo! Today Module user click log datasets with 45,811,883 user visits to the Today Module during the first 10 days in May 2009 \citep{shi2018massive}.

Geographical and climate problems always involve big data as well. \citet{guhaniyogi2017divide} considered the problem of capturing the spatial trends and characterizing the uncertainties in the sea surface temperature data in the west coast of mainland U.S., Canada, and Alaska from NODC World Ocean Database (www.nodc.noaa.gov/OC5/WOD/pr$\_ $wod.html). \citet{liang2013resampling} analyzed more than 100 year  data from the National Climatic Data Center from 1895 to 1997 (http://www.image.ucar.edu/GSP/Data/US.monthly.met).

Several publicly available movie scoring and music prediction data sets have been analyzed with divide-and-conquer approaches. \citet{tang2018learning} examined the MovieLens Data which is a popular public movie rating data set containing 20,000,263 movie ratings by 138,493 users of 27,278 movies from 1995 to 2015.  \citet{meng2017effective} and \citet{zhang2015divide} applied the divide-and-conquer approach to the Million Song Dataset (http://labrosa.ee.columbia.edu/millionsong/) which contains 515,345 songs with their years of release as the response.

\section{Discussions}
The divide-and-conquer approach is a general framework and it has been implemented in various models. Theoretical and numerical results demonstrate that the divide-and-conquer approach works well for big datasets. In many models where a simple average or weighted average is used, the combined results show the same efficiency as the results obtained by analyzing the entire dataset altogether. In more complex models such as Cox regression models, even the divide-and-conquer approach may not reduce  computational burden and time enough for practical use. An enhanced divide-and-conquer approach which includes linearization of original problem and one-step update strategy is utilized and demonstrates excellent performance. This is further extended to a multi-round divide-and-conquer framework. In addition, the combining step can be viewed as an optimization problem for certain loss function with regard to inferences from subsets. When a non-differentiable loss function is used, it can lead to median-based combining approaches.

One big challenge for the divide-and-conquer approach is how to choose $K$, the number of subsets. The choice of $K$  has been discussed in different models and the requirement of $K$ depends on the model as well as the rate of number of parameters. Several authors, e.g. \citet{tang2016method}, provide practical suggestions on the selection of $K$. However, a universal investigation and guidance would further improve the understanding and implementation of the divide-and-conquer approach.
The multi-round divide-and-conquer framework relaxes the requirement on the number of subsets $K$, which can be at the same order of total sample size $n$. Though the computational time can increase with the number of iterations, \citet{jordan2019communication} show that $O(\log{n}/\log(n/K))$ iterations would be sufficient.

\section{Acknowledgment} 

The authors wish to thank the editor and reviewer for their constructive comments and suggestions. The work is supported in part by US NSF grants DMS1737857, DMS1812048, DMS2015373 and DMS2027855. 

\bibliographystyle{chicago}

\bibliography{reference}

\end{document}